\documentclass{article}
\usepackage[utf8]{inputenc}
\usepackage{amsmath}
\usepackage{amssymb}
\usepackage{graphicx}
\usepackage{url}
\usepackage{makecell}
\graphicspath{ {./images/} }

     \usepackage[preprint]{neurips_2019}

\usepackage[utf8]{inputenc} % allow utf-8 input
\usepackage[T1]{fontenc}    % use 8-bit T1 fonts
\usepackage{hyperref}       % hyperlinks
\usepackage{url}            % simple URL typesetting
\usepackage{booktabs}       % professional-quality tables
\usepackage{amsfonts}       % blackboard math symbols
\usepackage{nicefrac}       % compact symbols for 1/2, etc.
\usepackage{microtype}      % microtypography

\usepackage{listings}
\usepackage{xcolor}

\definecolor{codegreen}{rgb}{0,0.6,0}
\definecolor{codegray}{rgb}{0.5,0.5,0.5}
\definecolor{codepurple}{rgb}{0.58,0,0.82}
\definecolor{backcolour}{rgb}{0.95,0.95,0.92}

\lstdefinestyle{mystyle}{
    backgroundcolor=\color{backcolour},   
    commentstyle=\color{codegreen},
    keywordstyle=\color{magenta},
    numberstyle=\tiny\color{codegray},
    stringstyle=\color{codepurple},
    basicstyle=\ttfamily\footnotesize,
    breakatwhitespace=false,         
    breaklines=true,                 
    captionpos=b,                    
    keepspaces=false,                 
    numbers=left,                    
    numbersep=5pt,                  
    showspaces=false,                
    showstringspaces=false,
    showtabs=false,                  
    tabsize=2
}

\title{Improving Deep Tabular Learning}

\author{
  Sivan Sarafian, Yehudit Aperstein \\ Intelligent Systems, Afeka Academic College of Engineering, Tel Aviv, Israel
}

\begin{document}

\maketitle

\begin{abstract}
	Tabular data remain a dominant form of real-world information but pose persistent challenges for deep learning due to heterogeneous feature types, lack of natural structure, and limited label-preserving augmentations. As a result, ensemble models based on decision trees continue to dominate benchmark leaderboards. In this work, we introduce RuleNet, a transformer-based architecture specifically designed for deep tabular learning. RuleNet incorporates learnable rule embeddings in a decoder, a piecewise linear quantile projection for numerical features, and feature masking ensembles for robustness and uncertainty estimation. Evaluated on eight benchmark datasets, RuleNet matches or surpasses state-of-the-art tree-based methods in most cases while remaining computationally efficient, offering a practical neural alternative for tabular prediction tasks.
\end{abstract}

% keywords can be removed

\section{Introduction}

Tabular Learning is a supervised learning paradigm represented as \(y = f(x)\), where \(y\) typically denotes a label or scalar value and \(x\) comprises discrete scalar features originating from real-world observations. For instance, \(x\) might represent an individual's medical records, while \(y\) could depict their projected height. Although tabular data may be predominant outside academia, many Machine Learning (ML) studies have prioritized more structured datasets like images and audio. This combination of diminished academic focus and inherent challenges in tabular data has often resulted in sophisticated deep learning algorithms underperforming when compared to traditional techniques like Decision Trees or state-of-the-art methods like Gradient-Boosting Decision Trees (GBDT).

Several complexities make tabular data challenging for general deep learning architectures, such as multilayer perceptrons. Primarily, there is no inherent knowledge of data structure in tabular data, unlike images or time series. In images, for example, it is generally understood that each pixel is connected to its neighboring pixels. Similarly, in time series, every data point has a sequential link, while in graph-based problems, an existing graph structure provides a natural understanding of relationships. A distinct domain structure can guide researchers in developing neural models with a strong inductive bias that captures the natural information flow in the problem. For example, recurrent neural networks \cite{hochreiter1997long} are suited to model the sequential information link which exists in time series problems.

This absence of structural understanding leads to difficulties in deriving superior representations for tabular data. As a result, neural network architectures tailored for tabular learning have been less advanced than those for other domains, like vision or audio. Furthermore, this lack of inherent structure means limited availability of effective augmentations, i.e. essential data transformations that maintain or predictably alter ground-truth labels. Such augmentations are vital in deep learning, mitigating overfitting and introducing beneficial noise, culminating in more robust models. Some learning frameworks, especially within the self-supervised learning (SSL) domain, rely heavily on these augmentations. Consequently, tabular data's paucity of augmentations hampers the extraction of meaningful representations from unlabeled data, a routine occurrence in other deep learning domains.

Another inherent challenge in tabular data is the amalgamation of varied data types. Broadly, features are either categorical (e.g., country, marital status) or numerical (e.g., weight, age). Yet, certain features can be interpreted both ways. The event time, for instance, might be viewed as a numerical feature—seconds since a reference—or as a categorical feature, such as day-of-the-year or a combination of related categorical and numerical features. Determining the feature type and its representation, generally an engineering choice, can significantly impact the predictor's success. Combining these varied data types into a cohesive model presents another challenge.

The scarcity of effective deep learning models for tabular data has elevated non-neural-network-based models, with Gradient-Boosting-Decision-Tree models (GBDT) being particularly noteworthy. These, comprising an ensemble of Decision Trees enhanced by gradient boosting techniques, have prominent algorithms like XGBoost, LightGBM, and CatBoost demonstrating state-of-the-art or near state-of-the-art performance. Their inherent simplicity, versatility, and efficiency, juxtaposed against the typically resource-intensive training of deep neural networks, make them a preferred choice for many practitioners tackling tabular prediction tasks.

In this endeavor, we aim to confront the challenges of deep tabular learning, proposing a versatile algorithm suitable for real-world application by data science practitioners. Our ambition is to craft a nimble algorithm capable of rivaling GBDTs in both efficacy and efficiency. Owing to computational constraints, our focus will remain on models with a runtime up to one hour on a single Nvidia GPU, excluding exceptionally large datasets. We posit that these boundaries, emphasizing more resource-efficient models, will resonate with data scientists seeking ready-to-use solutions and data analysts often working without substantial computational resources.

\section{Literature Survey}

Classical methods for tabular learning, including shallow learning algorithms and decision tree models, have long been utilized in the field of machine learning due to their interpretability, simplicity, and computational efficiency. Among shallow algorithms, Support Vector Machines (SVMs) \cite{cortes1995support} are notable for their wide use in various tabular data tasks. SVMs, proposed by Cortes and Vapnik in 1995, are particularly well-suited for high-dimensional data and have demonstrated strong performance across a variety of tasks. Multi-Layer Perceptron (MLP), a fundamental artificial neural network model, has also been used as an effective tabular data classifier despite its somewhat higher complexity compared to SVMs. MLPs, which consist of at least three layers of nodes in a directed graph, can capture complex patterns and relationships in the data \cite{bishop1995neural}. Decision trees, another classical method, are favored for their interpretability and ability to handle both continuous and categorical data. Decision tree models like \cite{quinlan1986induction} and \cite{salzberg1994c4} not only perform well in tabular learning tasks but also serve as the basis for more complex ensemble methods such as Random Forests and Gradient Boosting Machines.

Gradient Boosting Decision Trees (GBDT) have gained prominence for their robustness and efficacy in tabular learning. First introduced in 2001, GBDT models work by building an ensemble of weak prediction models, typically decision trees, in a stage-wise fashion \cite{friedman2001greedy}. Over time, GBDT has seen several innovative extensions that improved upon its original performance and efficiency. Among them, XGBoost (eXtreme Gradient Boosting), introduced in 2016 \cite{chen2015xgboost}, is particularly notable. XGBoost stands out for its scalability, speed, and superior performance across a wide range of datasets, making it a go-to algorithm in many machine learning competitions. Another extension, LightGBM \cite{ke2017lightgbm}, developed by Microsoft, introduced several novel techniques such as Gradient-based One-Side Sampling (GOSS) and Exclusive Feature Bundling (EFB) for handling large-scale data and high-dimensional feature space. Lastly, CatBoost, proposed in 2018 \cite{prokhorenkova2018catboost}, has been recognized for handling categorical features directly with its innovative Ordered Boosting method.

Post the deep learning revolution, advanced neural architectures were proposed as promising contenders for learning on tabular data. Particularly, the introduction of Transformer-based models for tabular data marked a new era. Originally designed for natural language processing tasks, Transformers \cite{vaswani2017attention} have been adapted for tabular data with notable success. TabTransformer \cite{huang2020tabtransformer}, for instance, employs Transformer architecture to learn global feature interactions in tabular data, demonstrating significant performance improvements. Another model, SAINT, utilizes a combination of Transformer and MLP architectures to learn robust representations from tabular data \cite{somepalli2021saint}. The recently introduced TabNet \cite{arik2021tabnet} and RTDL \cite{gorishniy2021revisiting} are two of the more popular deep learning methods for tabular data, both claim superiority over GBDT algorithms. While these approaches pushed the boundaries of tabular learning and hold promise for future advancements in the field, other studies have still raised skeptical concerns on the validity of their performance and suggested that GBDT algorithms still hold the top places in the leaderboard table \cite{shwartz2022tabular}.

In the years that followed the deep learning revolution, the battle between GBDT models and deep learning architectures has continued to rage with more new and successful deep learning algorithms, on the one hand, \cite{gorishniy2022embeddings,rubachev2022revisiting,borisov2022deep,huang2012tabtransformer}, but some mounting empirical evidence of the superiority of GBDT, on the other hand, \cite{grinsztajn2022tree,fayaz2022deep}. Specifically \cite{grinsztajn2022tree} has shown that GBDT outperforms deep learning algorithms in small to medium-size datasets and suggested that the tendency for smooth solutions of neural-network-based algorithms hinder the prediction performance with respect to the more non-smooth and even non-continuous solutions that can be learned with GBDT. Acknowledging the advantages of GBDT methods, some researchers have shifted towards hybrid approaches, in the hope to get the best of both worlds \cite{popov2019neural,ke2018tabnn,ke2019deepgbm}.

More recently, with the introduction of Large Language Models \cite{chang2023survey} such as ChatGPT \cite{wu2023brief} there have been attempts to harness the knowledge acquired by these massive models in order to improve or to completely replace classical tabular models \cite{hegselmann2023tabllm}. To incorporate LLMs as part of the classification task, the tabular data can be presented as a series of key-value numbers or categories that the model can interpret their meaning with additional metadata serving as contextual guidance. Then, the models may be implemented in several ways. The most basic approach is to design prompts with Few-Shot examples and rely on the LLM ability to generalize and adapt to new tasks \cite{li2024cancergpt}. More advanced methods attempt to fine-tune the model with the labeled examples \cite{hegselmann2023tabllm}. 

Alongside LLMs a sibling concept has emerged: Foundation Models (FM) \cite{bommasani2021opportunities}. FMs are highly versatile very large models, pre-trained on massive amounts of a blend of labeled and unlabeled data which can be adopted with a few-shot learning or fine-tuning to a downstream task. This concept has found popularity in vision, audio, and of course text (i.e. LLMs). To generalize the concept of FMs to tabular data \cite{yang2023unitabe} have designed a protocol to serialize tabular data s.t. the FM model can digest any table with any size, then, it may be fine-tuned for several downstream related tasks such as filling out missing value and classification.

While these attempts to utilize the power of large models have demonstrated what can potentially look like the future of data science, they have still failed to provide state-of-the-art classifiers. Nevertheless, Retrieval Augmented Generation (RAG) \cite{lewis2020retrieval} which is another new concept related to the world of NLP and LLMs is the silver bullet in a newly state-of-the-art deep tabular model. RAG are algorithms that given an input element can retrieve similar or relevant elements from a large data corpus. \cite{gorishniytabr} have used the RAG concept to incorporate a K-NN (K-nearest neighbors) augmentation to every element that needs to be classified. The K-NN retrieves examples from the labeled set (data and label) that supplement the model with auxiliary data has proved to be vital for improving the classification score. The authors have somewhat surprisingly been able to show that by using this approach they no longer need large and complex transformer models and they opt to demonstrate their model with an old-fashioned feed-forward simple model. Nevertheless, they were able to show an impressive set of result claiming that they surpassed significantly state-of-the-art GBDT models.

As this survey demonstrates, the world of tabular data relentlessly evolves and changes and new approaches are constantly suggested to improve this seemingly simple task. However, the battles of the key concepts of numerical datum representation, augmentations and model architecture have yet to be decided. To conclude this chapter we refer the reader to the last large comparison between neural-based models and gradient-boosting models \cite{mcelfresh2024neural}. In this paper, there have been introduced new open datasets and a comprehensive study of the dataset features that are better addressed by GBDT models and neural-based models, while they have not analyzed recent up-to-date models such as \cite{gorishniytabr} which claim superiority over GBDT models, their general statement is that GBDT models still dominate the battlefield of tabular learning.

\section{Suggested Algorithm}

Our proposed solution will address 3 fundamental challenges in deep tabular learning: (1)  model architecture (2) input representation (2) and; (3) augmentation.

\subsection{Model Architecture: the RuleNet Transformer}

At the beginning of the deep-learning era, linear feed-forward models dominated the field of tabular learning. Highly praised and popular models such as AlexNet and ResNet which were developed and demonstrated on vision tasks, were not immediately borrowed and adopted to solve tabular tasks, nor they were broadly used in other fields like audio and text. However, since the introduction of the Transformer architecture to the NLP world, the community has started to create more and more general purposed models that are based on the original concept of multi-head attention. Since then, transformers have found their place in NLP, vision, audio, and graph-based tasks. 

In the field of tabular learning, there have been a few attempts to incorporate transformers, or transformer concepts such as the multi-head attention \cite{gorishniy2021revisiting,arik2021tabnet}. Unfortunately, these architectures usually took advantage only of the encoder part of the original transformer, the one that translates the initial feature representation into an encoded set of features that can then be used to predict labels. One problem with this approach is that the size of the encoder output is bounded by the number of input features and cannot be determined as a system hyperparameter. This means that in a dataset with a small number of features, the network representation power can be potentially too weak to model the prediction task, and on the other hand, in a dataset with a large number of input features, the required computational demand will be too high which will have to be compensated in reducing the embedding size or the number of hidden layers and again will result in a non-optimal configuration. 

We seek to find a suited model for tabular learning in that its dependence on the input size is reduced. To that end, we suggest incorporating into the architecture the full transformer stack which includes a decoder on top of the encoder output. The role of the original decoder is to generate a new sequence of tokens which is dependent on the encoded input sequence. The size of the decoded output can be arbitrarily large and it is not restricted to the original input size.

To further develop our proposal let us first explain the general architecture of the transformer's encoder in the case of tabular learning. Transformers were originally developed for the field of natural language processing where each word or sub-word (named token) is represented by an embedding vector of a size \textit{emb\_dim}. Therefore, in the encoder, there are $M$ embedding vectors that represent the input sequence. These translate to $M$ encoded vectors of the same size at the encoder output and later they are combined with $N$ embedding vectors that represent the decoded sequence. Together they are processed in the decoder to yield the next word of the decoded sequence.

The analogous of this model in the tabular learning world is that each input feature is considered as a \textit{word} and is \textit{expanded} into an \textit{emb\_dim} embedding vector. In previous models, the encoded features and the output of the transformer's encoder were aggregated and used to predict the output label without any analogous decoder concept. We suggest using a decoder, which in the context of NLP is used to hold the decoded output sequence, and treating the decoder input as a set of predefined \textit{rules} that interact with the encoded features to generate a decoded rule prediction. The set of initial rules $r = {\textbf{r}_i}_{i=0}^n$ will be comprised of a set of $n$ vectors with each vector having a size of the transformer embedding size \textit{emb\_dim}. The set of rules will be learnable and task-dependent, but not sample-dependent. tensor, s.t. for each sample from the dataset the set of rules will be fixed.

We may conceptually associate each rule with the question of whether some condition was met in the given row of data, somewhat similar to the concept of splitting decisions in decision trees or to the concept of learnable filters in Convolution Neural Networks. In this case, however, the number of rules can be defined as an important hyperparameter that can be explored where more rules lead to a more complex model that can potentially fit harder problems and fewer rules can lead to smaller lean models that potentially, learn faster and suffer less over-fitting. We term our holistic unified model as the RuleNet.

Analyzing the computational budget of the Transformer's encoder-only architecture versus our suggested RuleNet, shows that the added decoder can actually reduce the total neural network FLOPS (floating point operations) required to solve the task. The number of FLOPs in the transformer is proportional to the number of key-value interactions (i.e. the number of non-zero elements in the attention matrix) times the squared size of the hidden layer $e_d^2$, times the network depth (i.e. the number of transformer layers). For a size of $M$ features and encoder size of $L_e$ layers, the total FLOPS is $O(M^2 \dot L_e \dot e_d^2)$ as there are $M$ features which interact with each other. For the decoder on the other hand, with input size of $M$ encoded features and $N$ rules and size of $L_d$ layers, the total FLOPDS is $O((M+N)N\dot L_d \dot e_d^2)$ as there are $M+N$ input \textit{keys} that interact with only $N$ output queries. Therefore, the total FLOPS for the encoder-decoder stack is\footnote{While the size of the hidden layer can be cancelled out when comparing the number of FLOPs in the decoder and the encoder, it is important figure when comparing with the total number of weights in the network}:
\begin{equation}
    FLOPS(encoder+decoder) = O(M^2 \times L_e \times e_d^2 + (M+N)N\times L_d \times e_d^2)
\end{equation}

On the other hand, the total number of weights in the transformer is comprised of the number of embedding vectors plus the size of the key-query-value networks in each transformer stack which is proportional to the \textit{emb\_dim} square. For the encoder-decoder stack, it is equal to $O(M \cdot e_d + N \cdot e_d + e_d^2 \times L_e + e_d^2 \times L_d)$ where $e_d$ in our case is both the embedding dimension size and the hidden vector size.\footnote{In practice while they may not necessarily be equal their size should correlate s.t. larger and smaller models increase and decrease both numbers accordingly.} In practice, we want to learn a model with a sufficient capacity where the capacity is proportional to the total number of weights. Therefore, without a decoder, in order to increase the model's capacity, the number of encoder layers $L_e$ must be increased. However, this can result in an overwhelming number of FLOPS due to the \textbf{square dependency} of FLOPS in the input dimension size $FLOPS \sim M^2$. Note that one cannot simply increase $e_d$ as it will similarly increase the hidden layer size as it will equally increase the number of FLOPs. Therefore, in the case of encoder-only in our stack, the optimal model for a large number of features quickly consumes the computational capacity.

Incorporating a decoder into the network helps in reducing the dependency of the model capacity on the number of encoder layers, s.t. the total number of FLOPs can be reduced while maintaining a sufficient number of weights. We can design a model with a fixed number of rules and delegate most of the model capacity to the decoder part \footnote{where in the extreme case of a very large number of input features we can eliminate the encoder entirely.} such that the total number of FLOPS is maintained and depends only on the first power of $M$, i.e. $FLOPS \sim M$.

Decoder also helps in the case of a very small number of input features $M$. In this case, while the computational budget is controlled, in order to generate sufficient capacity to learn large datasets without a decoder, one needs to increase the number of encoder layers $L_e$ or the size of the hidden vector dimension $e_d$. Both options may result in an unbalanced network (too deep neural-net or too wide neural-net). Using the decoder, we can effectively simulate input to the neural net, as the rules can be viewed as an auxiliary input that aggregates data from the entire dataset. Therefore, setting the number of rules $N$ is an effective tool to stabilize the size of the input to the network and balance the overall network architecture to avoid too deep a stack of layers or too wide a network.

\begin{figure}[h]
    \centering
    \includegraphics[width=.75\textwidth]{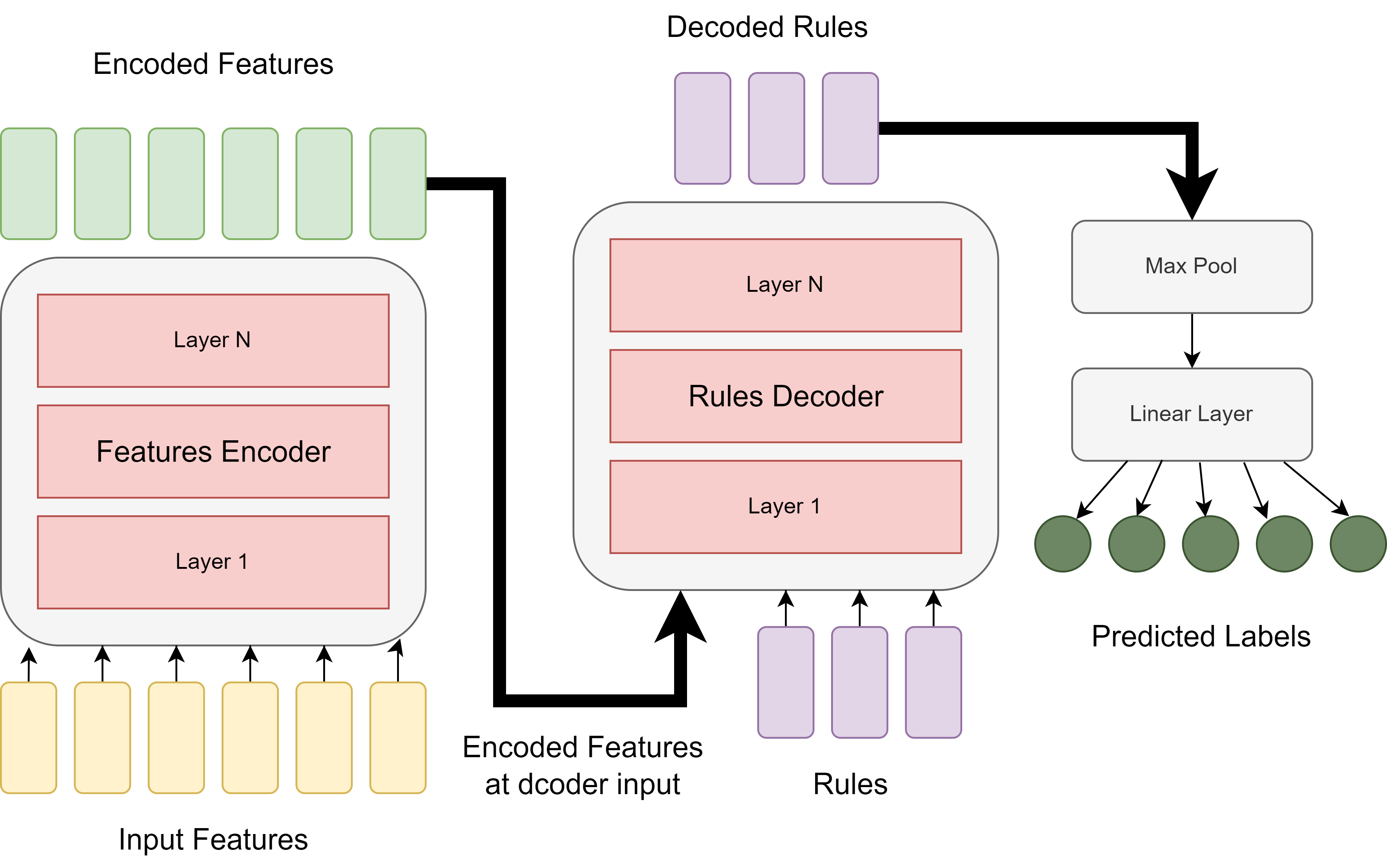}
    \caption{The RuleNet Transformer}
    \label{fig:rulenet}
\end{figure}

In Fig. \ref{fig:rulenet} the RuleNet Transformer architecture is illustrated. The input features play the role of input token to the encoder and the rules play the role of output token to the decoder. The representation of the input features is described in Sec. \ref{sec:Input representation} and the representation of each rule is a learnable embedding vector that is not input dependant. As in the original transformer, the encoded input tokens (i.e. the encoded features) are passed to the decoder. However, unlike the original transformer, we do not use causal attention mask in the decoder and each rule has access to all the encoded features as well as all other rules. In addition unlike the original transformer we do not use positional encoding, every feature and rule has its own independent representation and there is no particular order in the set of features and rules so there is no need for extra positional encoding. At the output of the transformer, a MaxPool layer aggregates the rule stack of size $N\times e_d$ to a size of $e_d$ and then a linear layer outputs either a single scalar (in regression problems) or a $N_c$ logits in classification problems where $N_c$ is the number of classes.

\subsection{Input representation: Piecewise Linear Quantile Projection}
\label{sec:Input representation}

One crucial part of transformer architecture is the necessity of representing each input as an embedding vector. In our task, it means that we need to transform each input feature into a learnable vector. In the case of categorical features, this is actually a relatively easy task with not so many possible alternatives. Like token words, each category is treated as a different token and is associated with a different, independent, learnable embedding vector. However, in the case of numerical features, the question of how to transform a scalar value into a meaningful embedding vector does not have so much obvious solutions. To suggest our solution, it is better to split the problem into two consecutive parts:
\begin{enumerate}
    \item \textbf{Prepossessing:} In this step a scalar value is processed with a predefined, non-learnable transformation. Usually, this transformation is based on some statistics, measured over the training set. For example, Z-score normalization or robust scaler are two popular normalization methods \cite{sklearn_api}. In these methods, the output of the preprocessing part is a scaled numerical value. On the other hand, Quantile transformer can also be used to preprocess the numerical value and transform it to the $[0,1]$ range. However, unlike other scalers, Quantile transformer can map numerical values to discrete and equally populated bins, where each bin is associated with a different quantile value. For example, if we use 100 quantiles uniformly spaced between $[0-1]$ we obtain the precentiles of the data s.t. each bin represents a different percentile and in each bin there are exactly $1/100$ samples of the total training data. The advantage of this discrete representation is that we transform a numerical value into a categorical set of tokens, where we can then apply the same embedding representation as done with categorical features. Note that one obvious drawback of the transformation into categorical values is that the inductive relationship between two closed values is broken in such a model and this reduces the generalization capacity of the model.
    
    \item \textbf{Expansion:} after the preprocessing part we need to expand the scalar value into an embedding vector. This part is usually done with a learnable set of weights but it is not obligatory. For example, the most naive approach can be to assign each token with a different predefined random vector. Alternatively, one can use harmonic functions to represent a value $\textbf{e}_x = [\cos(\omega_i x),\sin(\omega_i x),...]_{i=0,..,N-1}$. However, usually, learnable expansion is preferred to be able to adjust to the specific problem at hand. In \cite{gorishniy2021revisiting} the authors decided to use a linear learnable transformation $\textbf{e}_x = \textbf{w} \times x$ to expand a scalar value into a vector and in \cite{arik2021tabnet} the authors used a deeper network. Usually, a deep network is required to generate a composite, non-linear transformation. However, too-deep transformation can be unnecessarily computationally heavy as it is done for each different scalar numerical feature.
\end{enumerate}

Our suggested solution is based on a composition of two approaches: (1) quantile transformation from numerical values to categorical values and; (2) piecewise linear transformation of the interpolated quantile value. For a given numerical scalar feature $x^j$, where $j$ is the feature name, we first assign two values: (1) the discrete quantile index $i$ and; (2) the linear mapping of the value in the range of the two consecutive quantiles:
\begin{equation}
    f_x^j = \frac{x^j - q_i^j}{q_{i+1}^j - q_i^j}
\end{equation}
for each quantile index, we assign a different embedding vector but the total representation of $x^j$ is the piecewise linear transformation of $f_x^j$
\begin{equation}
    \textbf{e}_x^j = (1-f_x^j) \textbf{e}_{q_i^j} + f_x^j \textbf{e}_{q_{i+1}^j}.
\end{equation}
There are a few advantages of this representation over other schemes. First, unlike linear mapping, it can represent more complex and non-linear correlations between input features and output predictions. Second, it is much more lightweight in terms of both weights and computing than a feed-forward network (stacked layers of linear mapping and non-linear activations). 
 Third, unlike the pure numerical to categorical quantile transformation, it is a continuous transformation that 
 maintains the correlation between two close values. The last advantage of our approach is related to our suggested augmentation proposal, which will be outlined in what follows.

\subsection{Augmentation: features masking ensembles}

As previously explained, label-preserving augmentations in tabular data may seem like a white whale, as it is hard to universally define a transformation that preserves the label without first learning a model (such as image rotations for image classification tasks). However, augmentations like image crop which are often used in vision tasks, have an equivalent in tabular data which is feature masking. When applying a feature masking augmentation, the network is required to learn from a masked set of features where when a feature is masked it is replaced by a masked token. The difficulty is to encode the masked token into an embedding vector in order to preserve the same set of embedding vectors at the transformer entry. 

Again, for categorical features, it is a relatively easy task. We set a unique \textit{MASKED} token in our embedding list and when a feature is masked its embedding vector is replaced by the embedding vector of the \textit{MASKED} token. However, it is not so obvious that the same approach is possible for some numerical representation methods. For example, for the linear transform representation, what would be the value that replaces a masked feature? its mean value? zero?. Nevertheless, when we transform a numerical feature into a categorical token with the quantile transform we can apply the same masking method and use a MASKED token to replace masked numerical features. Therefore, our overall numerical feature representation is
\begin{equation}
    \textbf{e}_x^j = \begin{cases}
        \textbf{e}_{MASKED}^j & \varepsilon \leq p_{mask} \\
    (1-f_x^j) \textbf{e}_{q_i^j} + f_x^j \textbf{e}_{q_{i+1}^j} & \varepsilon > p_{mask}
    \end{cases}
\end{equation}
where $p_{mask}$ is the masking probability and $\varepsilon\sim U(0,1)$. In this case, the masked numerical feature behaves similarly to its categorical counterpart and it is also replaced with a learnable embedding vector of the \textit{MASKED} token. 

The last ingredient in our recipe is the \textit{ensemble} of masking. Usually, in inference time the augmentations are turned off in favor of better accuracy of the model. However, there is an inherent and sometimes overlooked disadvantage in turning off the augmentations in inference time. By doing so, the statistics of the test set diverge from the train set which was used to train the model. The model never saw a complete set of features during training and therefore, going out-of-distribution (OOD) can sometimes potentially hamper the performance instead of boosting it. An alternative approach that avoids from risking in OOD can be to generate a prediction based on an average of \textit{weak} predictors where each weak predictor is the model output to a different augmentation of the data
\begin{equation}
    y_{\theta}(x) = \frac{1}{K}\sum_i y_{\theta}(x^i_{AUG})
\end{equation}
This approach requires rolling out $K$ predictions out of the model to yield a single result, hence it is more computationally heavy. However, if the computational budget is acceptable, it is a much more appealing approach than the naive "turn augmentation off" approach as it bears another unique advantage: it provides a built-in uncertainty estimation of the model. The uncertainty of the model can be defined as the standard deviation of the prediction
\begin{equation}
    \sigma_{y_{\theta}}(x) = STD([y_{\theta}(x^i_{AUG})|i=1,...,K])
\end{equation}
and it can be used to assess whether the model is certain in its prediction (a small standard deviation corresponds to higher certainty). We note that the ensemble approach can be expanded to include all the stochastic elements in the model besides masking, i.e. also batch normalization and dropout layers. In our case, we avoid adding batch normalization in our model, but we do add dropout layers in the transformer and in the linear output layer. Therefore, we add also the dropout rolling to our definition of the ensemble of models.

\section{Experimental Setup and Datasets}

As explained earlier, we designate our implementation and analysis to real-world problems and algorithms that can be executed on relatively lightweight hardware in a relatively short time. Therefore, we restrict our experiments to a training process that runs no more than several hours on a single GPU (NVIDIA RTX 3090Ti). This choice is both due to limited access to compute resources and also since many data-science practitioners, would find such lightweight algorithms much more appealing than heavy-duty algorithms which are more complex to implement, require more GPU devices, and run for longer time.

For the datasets, we follow the baselines in \cite{gorishniy2021revisiting} and evaluate both classification and regression problems. For regression problems, we use the Mean Squared Error (MSE) of the z-score normalized value for the loss function
\begin{equation}
    \mathcal{L}(\theta) = \sum_{i} \left( \hat{y}_{\theta}(x_i) -  \frac{y_i - \mu_y}{\sigma_y + \varepsilon} \right)^2
\end{equation}
and we evaluate the ROOT-MSE (RMSE) as the objective.

In total, we choose 8 out of 11 datasets evaluated in \cite{gorishniy2021revisiting}, we omit the larger datasets with a large set of features as they require much more training resources. 
\begin{table}[ht!]
    \centering
    \begin{tabular}{l|c|c|c|c|c|c|c}
    \hline
        Name & Ref. & Abbr. & Objects & N-features & Cat-features & Metric & \#-Classes \\
    \hline\hline
        California Housing & \cite{pace1997sparse} & CA & 20640 & 8 & 0 & RMSE & - \\
        Adult & \cite{kohavi1996scaling} & AD & 48842 & 6 & 8 & Acc & 2 \\
        Helena & \cite{guyon2019analysis} & HE & 65196 & 27 & 0 & Acc & 100 \\
        Jannis & \cite{guyon2019analysis} & JA & 83733 & 54 & 0 & Acc & 4 \\
        Higgs & \cite{baldi2015enhanced} & HI & 98050 & 28 & 0 & Acc & 2 \\
        ALOI & \cite{geusebroek2005amsterdam} & AL & 10800 & 128 & 0 & Acc & 1000 \\
        Year & \cite{bertin2011million} & YE & 515345 & 90 & 0 & RMSE & - \\
        CoverType & \cite{blackard1999comparative} & CO & 581012 & 54 & 0 & Acc & 7 \\
    \hline
    \end{tabular}
    \hspace{5mm}
    \caption{Dataset details}
    \label{tab:datasets}
\end{table}

\subsection{Preliminary Results}
\label{sec:Preliminary Results}

To quickly test our approach we conducted a simple experiment where we manually tuned the RuleNet algorithm alongside a catboost baseline on a single dataset and then we let the algorithms learn the rest of the datasets without further hyperparameter optimizations. To explore and tune the algorithm's hyperparameters we chose the CoveType dataset which we have previous experience with. In our experiments we always use the GeLU activation \cite{hendrycks2016gaussian}, AdamW Optimizer \cite{loshchilov2017decoupled}, One cycle scheduler \cite{smith2019super} with fixed 100 learning epochs, and the default weights initialization methods provided by PyTorch \cite{NEURIPS2019_9015}. The rest of the hyperparameters including the learning rate, size of the transformer network, number of quantiles, and more, can potentially be tuned in a hyperparameter optimization routine, depending on the available budget. %The set of initial hyperparameters which we found that best fit the CovType dataset is listed in Table \ref{tab:hp}.

Table \ref{tab:results} summarizes all the results of our RuleNet algorithm alongside the Catboost baseline which we trained and the State-Of-The-Art algorithms their results, as were reported in \cite{gorishniy2021revisiting}. Even with only minimal manual tuning, RuleNet showed high competitiveness and scored SOTA performance in two out of our 8 datasets: CovType and ALOI. As we tuned our hyperparameters on the CovType dataset, it is expected that the RuleNet results should be strong but the fact that on a first run without any further tunning it yielded SOTA results on ALOI is indeed impressive. On the other hand, on smaller datasets, it still requires more tuning iterations but its results are somewhat on par with un-tuned Catboost model. As can be seen from Sec. \ref{hpo_optimization}, while small datasets are known to be a tough hurdle for deep neural network algorithms, hyperparameter optimization can significantly improve the performance also in this task segment.

\begin{table}[ht!]
    \centering
    \resizebox{\textwidth}{!}{%
    \begin{tabular}{l|c|c|c|c|c|c}
    \hline
        Name & Objective & RuleNet & Catboost & Catboost training-time [Sec] &  \makecell{SOTA \\ (paper-results)} &  \makecell{SOTA \\ algorithm}\\
    \hline\hline
        CA            & RMSE  &  0.449   &  0.476   &   232        &  0.43 &  Catboost \\
        AD            & Acc   & 0.861    &  0.862   &   4480       &  0.874 & XGBoost \\
        HE            & Acc   & 0.378    &  0.331   &   6520       &  0.396 & ResNet \\
        JA            & Acc   &  0.718   &  0.706   &   1390       &  0.732 & FT-Transformer \\
        HI            & Acc   & 0.716    &  0.722   &   868        &  0.729 & FT-Tranformer \\
        \textbf{AL}   & Acc   &  0.969   &  0.941   &   107,640    &  0.963 & ResNet  \\
        YE            & RMSE  &  8.771   &  8.74    &   528        &  8.784 & NODE \\
        \textbf{CO}   & Acc   & 0.9763   &  0.965   &   1817       &  0.973 & FT-Transformer \\
    \hline
    \end{tabular}
    }
    \hspace{5mm}
    \caption{Preliminary Results}
    \label{tab:results}
\end{table}

\subsection{Hyperparameter Optimization}
\label{hpo_optimization}

To fairly compare RuleNet with the literature SOTA, we need to rigorously conduct Hyperparemeter Optimization (HPO) research s.t. we use the validation sub-set as a set to optimize the best hyperparameter set for each dataset \cite{gorishniy2021revisiting}. The validation set is also effectively used to find the best early stopping point to avoid overfitting the data. To that end Table \ref{tab:hpo_parameters} details all the hyperparameters which we select to optimize. We implemented and tested the HPO process using Ray-Tune \cite{liaw2018tune}, a Python package that can effectively manage multiple concurrent runs. As the optimization algorithm, we selected the OptunaSearch algorithm which is based on the Optuna package \cite{akiba2019optuna} implementation of the Tree-structured Parzen Estimator algorithm \cite{ozaki2020multiobjective} and the ASHA scheduler \cite{li2018massively} to prune unsuccessful trials before they terminate.

\begin{table}[ht!]
    \resizebox{\textwidth}{!}{\begin{tabular}{ccccc}
    \hline
        Hyperparameter & Type & Min Value & Max Value & Notes\\
    \hline
    \hline
        Batch Size & Categorical & $BS/4$ & $2BS$ & $BS$ is the optimal batch size in \cite{gorishniy2021revisiting} for each dataset\\
        learning rate (dense)   & numerical     & $10^{-4}$     & $10^{-2}$     & controls all weights besides the embedding block \\
        learning rate (sparse)  & numerical     & $10^{-3}$     & $10^{-1}$     & controls the embedding block learning rate \\
        $N$ rules & Categorical & 64 & 256 & Number of decoded rules \\
        $n$ quantiles & Categorical & 2 & 100 & Number of quantiles where 2 is effectively no quantization\\
        embedding dim & Categorical & 64 & 256 & Size of the embedding vector\\
        Encoder Layers & Categorical & 1 & 8 & Number of encoder layers \\
        Decoder Layers & Categorical & 1 & 8 & Number of decoder layers \\
        Transformer heads & Categorical & 1 & 8 & Number of parallel heads (both encoder and decoder) \\
        Transformer hidden dim & Categorical & 128 & 512 & Latent vector size \\
        mask-rate & Numerical & 0 & 0.4 & Probability of masking the input features \\
        rule-mask-rate & Numerical & 0 & 0.4 & Probability of masking the rule features \\
        Transformer-Dropout & Numerical & 0 & 0.4 & Internal dropout of the transformer \\
        Dropout & Numerical & 0 & 0.5 & Dropout in the linear post transformer layer \\
        Label smoothing & Numerical & 0 & 0.4 & A smoothing regularizer for the loss function\\ 
\hline
    \end{tabular}}
    \hspace{5mm}
    \caption{Hyperparameters of RuleNet that were optimized}
    \label{tab:hpo_parameters}
\end{table}

Table \ref{tab:experiments} summarizes the results of hyperparameter tuning for each of our datasets. Due to the difference in size and training time, the number of total trials varies between datasets s.t. the total training time with 4 GPU cards (NVIDIA 3090Ti) was limited to around 48 hours. We can see that our algorithm beats the SOTA results in 4 out of 8 datasets and is on par with the SOTA results in 3 others. Overall, the total training time is less than 2 hours for each trial with the exception of the Covtype dataset which took approximately 3.5 hours due to the relatively large dataset and the large number of features.  We can observe that generally, the RuleNet excels in larger datasets and obtains on-par results for small datasets. This is an encouraging finding as before the HPO process our results in the small datasets such as CA and HE were slightly lower than the state-of-the-art. It is worth noting that HPO indeed improves the overall results in all the datasets however, the compute time might be too expensive for somewhat marginal improvement in real-life scenarios.

\begin{table}[ht!]
    \centering
    \begin{tabular}{ccccc}
    \hline
        Dataset & Score & Time & N-trials & best baseline\\
    \hline
    \hline
        CA & 0.433 & 346 & 562 & 0.43\\
        AD & 0.8693 & 1054 & 351 & 0.874\\
        HE & 0.3921 & 848 & 302 & 0.396\\
        JA & \textbf{0.7359} & 5590 & 351 & 0.732\\
        HI & 0.7266 & 2617 & 90 & 0.729\\
        AL & \textbf{ 0.9688} & 6480 & 161 & 0.963\\
        YE & \textbf{8.771} & 5848 & 139 & 8.784\\
        CO & \textbf{0.9763} & 12823 & 240 & 0.973\\
    \hline\\
    \end{tabular}

    \caption{HPO best results for each dataset}
    \label{tab:experiments}
\end{table}

\begin{figure}[ht!]
    \centering
    \includegraphics[width=1.\textwidth]{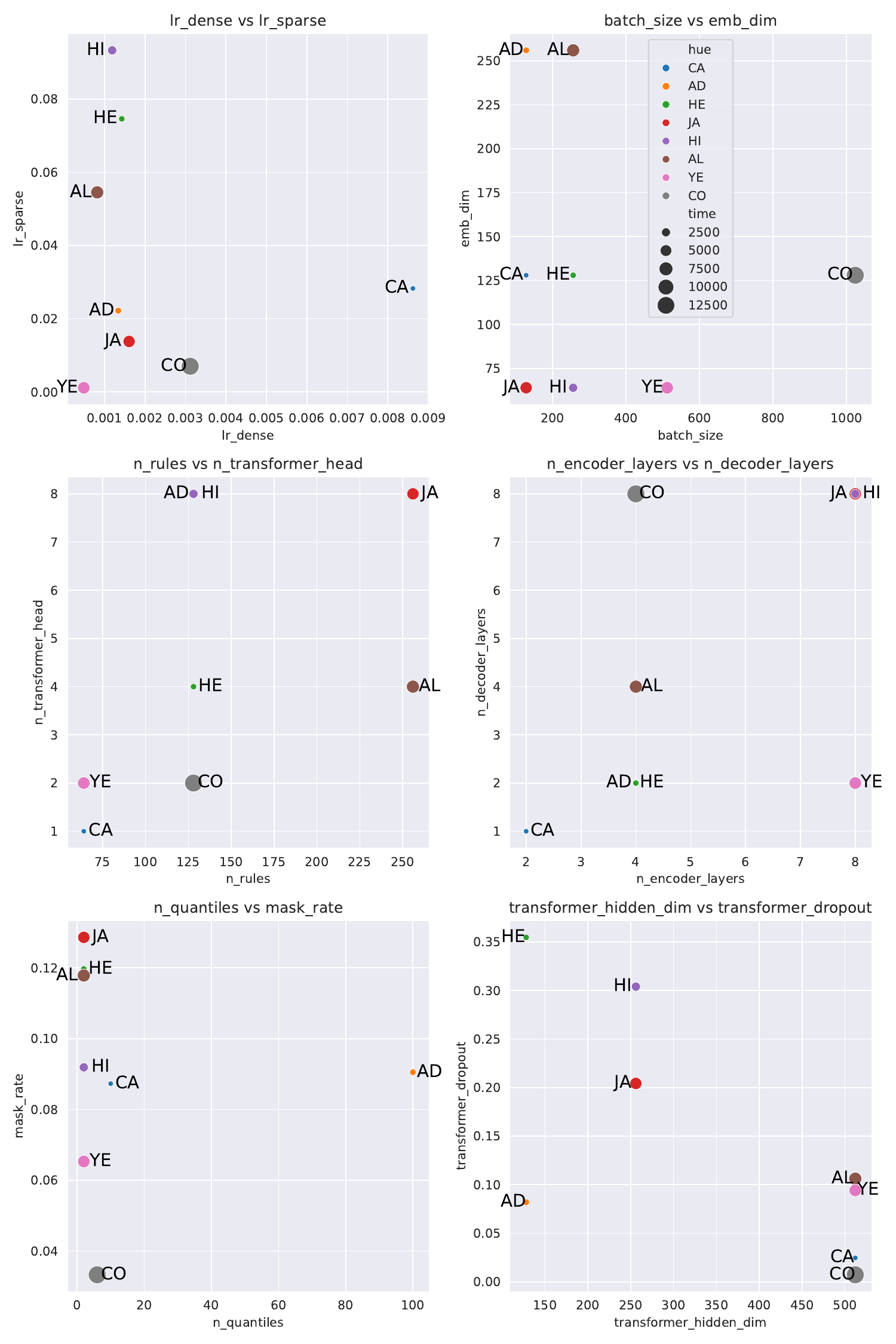}
    \caption{Hyperparameter distribution of the best trials}
    \label{fig:best_hpo_dist}
\end{figure}

In Figure \ref{fig:best_hpo_dist} we analyze the hyperparameter distribution of the best trials for each dataset. We take pairs of hyperparameters that have a close relationship together, e.g. dense and sparse learning rate and we plot the scatter plot of the best configuration for each dataset, the size of the maker corresponds to the size of the dataset which is usually a good proxy value to the complexity of the problem. We can observe that on general larger datasets require smaller learning rates yet the optimal sparse learning rate is relatively high with respect to the dense learning rate .This is because unlike the neural-network weights which contribute to the algorithm's output for any given input, the embedding weights of a specific category contribute to the algorithm only for inputs that contain this specific category, therefore effectively they exist in less training steps which must be compensated with a larger learning rate.

When comparing batch-size and embedding dimension, we can see the trend of larger batch size for larger datasets (as was also observed in \cite{gorishniy2021revisiting}). However, the correlation between dataset size and the embedding dimension is less obvious. Regarding the number of quantiles, 5 out of 8 datasets topped with $n_q=2$ this effectively translates to a linear transformation between the embedding that represents the 0th quantile and the embedding that represents the 100th quantile, therefore effectively while it uses the utilization scaling (i.e. the numerical input is transformed into a [0-1] feature) it does not use our piecewise linear quantile embedding. This is somewhat a disappointing finding, yet it can be seen that in 3 cases the quntile embedding did contribute to the overall score, especially in AD where the best results were obtain with $n_q=100$. On the other hand, the mask value almost uniformly distributed between 0.03 and 0.13 with mean value of around 0.9 which shows the effectiveness of adding the masking augmentation as part of the learning procedure.

With respect to the $N$-encoder layers versus $N$-decoder layer, we do not find a general rule of thumb. It looks like the decoder is indeed important to obtain the best results, yet the mixture of number of encoding and decoding layers is not precisely understood. To shed more light on these nuances, we turn to a sensitivity analysis of each of the hyperparameters.

\subsection{Sensitivity Analysis}

\begin{figure}[ht!]
    \centering
    \includegraphics[width=1.\textwidth]{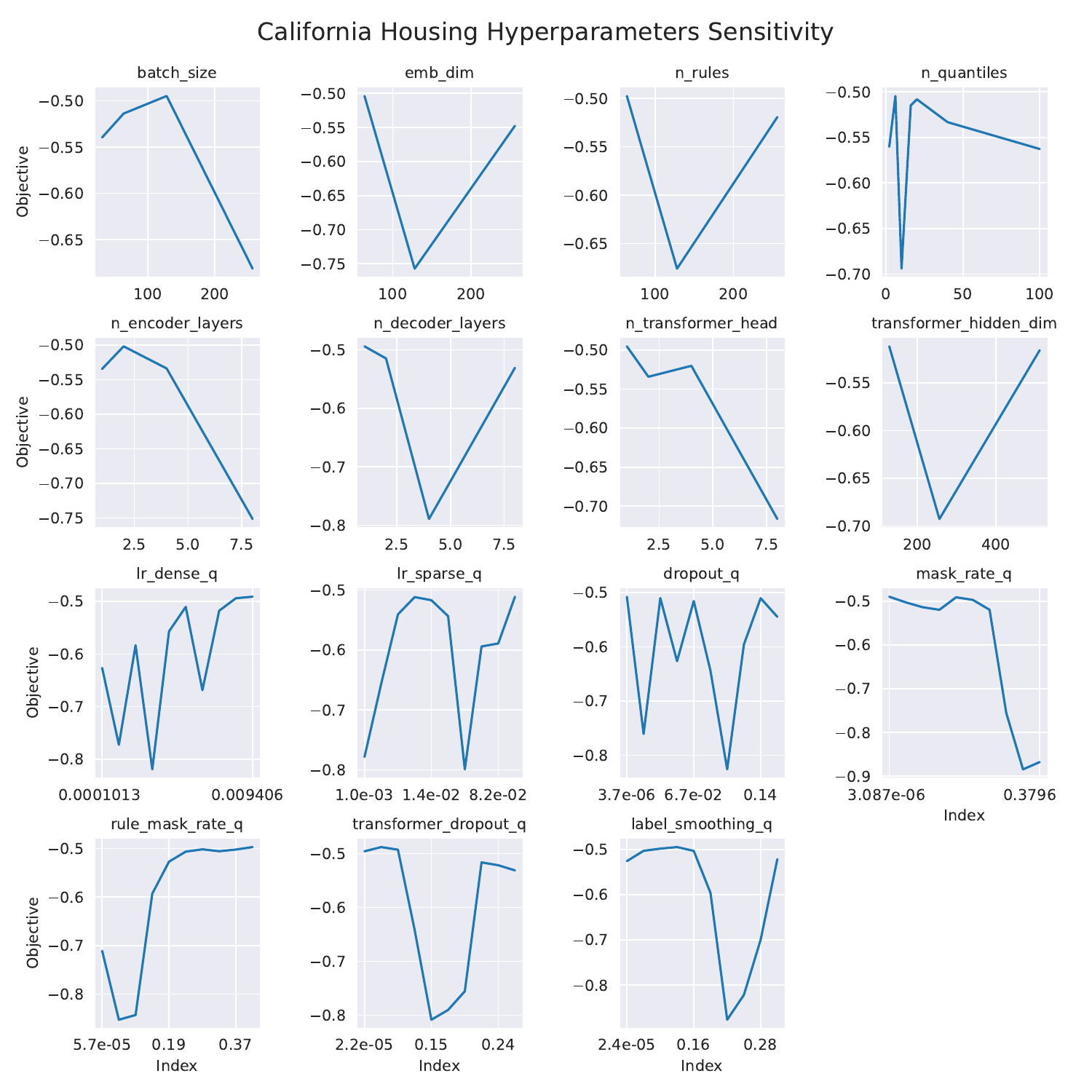}
    \caption{Hyperparameter optimization sensitivity analysis for the CA dataset}
    \label{fig:ca_hpo}
\end{figure}

\begin{figure}[ht!]
    \centering
    \includegraphics[width=1.\textwidth]{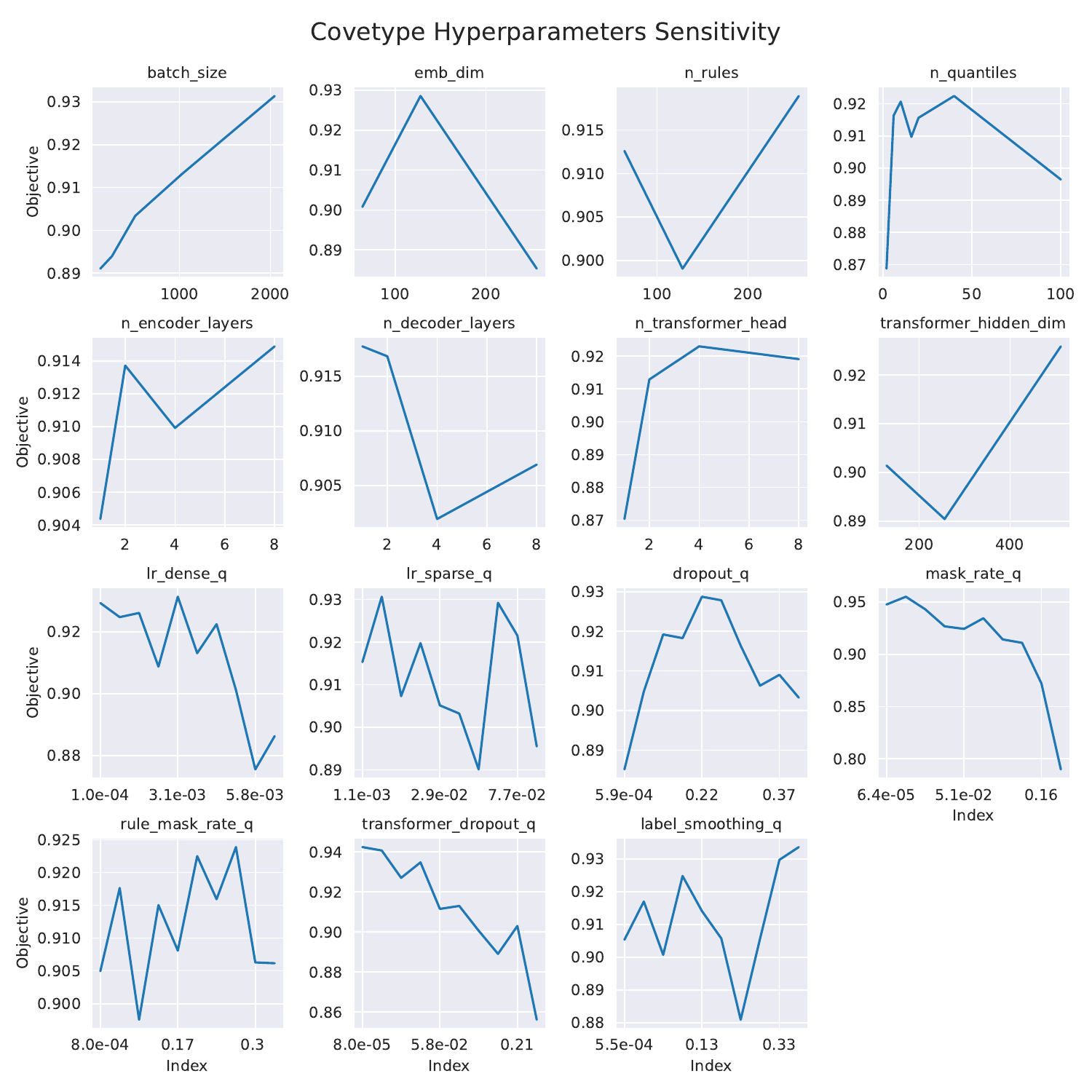}
    \caption{Hyperparameter optimization sensitivity analysis for the CO dataset}
    \label{fig:co_hpo}
\end{figure}

To check the marginal contribution of each hyperparameter value, we take the total set of trials (for each dataset) and group by the hyperparameter value and then average the score:
\begin{equation}
    V_{p_i}(\mu) = \frac{1}{N_\mu}s_j
\end{equation}
where $p_i$ is some hyperparameter like number of quantiles ($n_q$) $\mu$ is some hyperparameter value and $s_j$ is the score of an experiment where $p_i=\mu$ and $N_\mu$ is the total number of experiments where $p_i=\mu$. For numerical parameters, we first divide the hyperparameter values into a discrete set of buckets with the quantile transformation (effectively discretize the parameter after the run). We plot this analysis for two representative datasets: (1) California Housing (CA) in Figure \ref{fig:ca_hpo} and; (2) Covetype (CO), in Figure \ref{fig:co_hpo}. These datasets represent two extreme ends in our analysis of small and large datasets.

An important observation of the sensitivity analysis is the spread of the average objective value with respect to different hyperparameter values. A small spread points towards better robustness of the algorithm with respect to a specific parameter where larger spread indicates that the algorithm's sensitivity to an exact hyperparameter value is large. For example, analyzing Figure \ref{fig:co_hpo} we find that the algorithm is much more robust with respect to the $N$-encoder and $N$-decoder layer than with respect to the mask rate. In the case of higher mask rates, the score may significantly degrade. 

Another observation that can be derived from the sensitivity plots is whether our initial guess included the optimal value. If there is a concave shape with a dominant "sweet spot" we can fairly say that the range of the hyperparameter in our initial guess is correct, e.g. it is clear from the CO graph that the right amount of dropout is somewhat around 0.2. However if the trend does not include a clear inner maximum and the maximum is obtained either in the low or high end of the range, as it is the case in the batch size of CO, where we observe that larger and larger batches lead to better score, we can say that we need to rerun the test in order to find the absolute optimal parameters. We note that in the case of the batch size it is not possible due to computational constraints that limit our budget.

The analysis shows that many parameters are yet to obtain the desired sweet spot and the score can be potentially further improved with larger range of evaluation, yet many of these parameters (such as the size of hidden layers and number of layers) require increasing the computational budget which is outside the scope of this project. Comparing the sensitivity of the CA experiment parameters and the CO experiment parameters we can see that CA is much more sensitive to hyperparameters fine-tuning probably due to its relatively smaller size which requires much more focus on the regularization aspect and this another evident that RuleNet better fit larger datasets and more complex problems.

\subsection{Ablation Tests} 

To test the impact of each of our improvements, i.e. (1) masking augmentation (2) encoder-decoder architecture (3) piecewise linear quantile embedding, 
we run an ablation test on the Coveetype (CO) and California-Housing (CA) datasets with 4 different independent experiments\footnote{Note that in the previous section we have already found that only CO, CA and AD require $n_q > 2$ to achieve the best results, so we expect a negative contribution of the piecewise linear quantile embedding part in these datasets}:
\begin{enumerate}
    \item Full: a HPO process with all our improvements
    \item No-mask: a HPO process where the dropout/masking elements are disabled
    \item No-Dec: a HPO process where the decoder part is bypassed
    \item No-Quant: a HPO process where the quantization is reduced and fixed to 2.
\end{enumerate}
To restrict the budget, we capped the number of trials in each experiment to 100. The results in Table \ref{tab:ablation} show that for CO and CA all components of our RuleNet architectures are crucial to obtain state-of-the-art results. We find that in accordance with the previous section, the no-decoder experiments where second best which indicates the relative robustness of the overall algorithm to this parameter.

\begin{table}[ht!]
    \centering
    \begin{tabular}{ccc}
    \hline
        Dataset & Score  \\
    \hline
    \hline
        CA-full & \textbf{0.433} \\
        CA-no-decoder & 0.4419 \\
        CA-no-quantile & 0.45130 \\
        CA-no-masking & 0.456 \\
        CO-full & \textbf{0.9763} \\
        CO-no-decoder & 0.975000 \\
        CO-no-quantile & 0.974200 \\
        CO-no-masking & 0.9746 \\
         \hline\\
    \end{tabular}
    \caption{Ablation results on CA and CO datasets}
    \label{tab:ablation}
\end{table}

\section{Discussion}

Our project introduced a comprehensive and innovative approach to deep tabular learning. It addresses fundamental challenges associated with model architecture, input representation, and data augmentation. Our proposed algorithm, RuleNet, leverages an encoder-decoder transformer architecture, piecewise linear quantile embedding for numerical feature representation, and a feature masking ensemble technique for augmentation. These components collectively contribute to the model's robust performance across a diverse set of tabular datasets, showcasing the potential to achieve and surpass the state-of-the-art results which are primarily dominated by Gradient Boosting Decision Tree (GBDT) models. Nevertheless, in retrospect, we can analyze the importance, impact, and significance of each of our suggested improvements.

Starting with feature masking, our experiments have proven that this component has the highest contribution to the success of the overall algorithm. Importantly, unlike other parameters that change the network size or training time, this parameter does not change the algorithm training routine, therefore it is very easy to be optimized with HPO framework or some small manual search. Its single drawback is that, if used to generate an ensemble in inference time, it makes the inference heavier and slower which in some cases can pose a limitation.

Our second most promising direction is piecewise linear quantile embedding. While we found that it is not always a necessary complication, sometimes it is the key factor in obtaining state-of-the-art results. Like feature masking, it can be easily optimized manually or with HPO. Lastly, our suggested juxtaposed encoder-decoder architecture has also provided a positive impact on the overall score, yet its contribution is more marginal. We believe that when dealing with large datasets, our suggested method can be especially beneficial over encoder-only architecture, however, due to limited computational resources we could not evaluate this hypothesis in practice.

\section{Conclusions}

In this work, we developed a neural network based algorithm for learning supervised tasks such as classification and regression in the tabular data domain. Our algorithm aimed at tackling 3 different aspects in tabular data that are considered hard with respect to other domains, i.e.: (1) data representation via piecewise linear quantile embedding, (2) augmentation via feature masking and; (3) neural net architecture via encoder-decoder transformer architecture. We found through an extensive analysis of 8 datasets that our improvements indeed achieve state-of-the-art results or almost on-par results with the recent state-of-the-art. We found that on average all the 3 improvements contribute to the final score while sometimes our piecewise linear quantile embedding is inferior to simple quantile transformation multiplied by a single embedding vector. To conclude, we believe that all our suggested elements should be part of the swiss army knife of the data-scientist practitioner when he or she opt to solve a tabular task with neural networks.

%\newpage

\bibliographystyle{plain}
%\bibliography{mybib}

\begin{thebibliography}{10}

\bibitem{akiba2019optuna}
Takuya Akiba, Shotaro Sano, Toshihiko Yanase, Takeru Ohta, and Masanori Koyama.
\newblock Optuna: A next-generation hyperparameter optimization framework.
\newblock In {\em Proceedings of the 25th ACM SIGKDD international conference on knowledge discovery \& data mining}, pages 2623--2631, 2019.

\bibitem{arik2021tabnet}
Sercan~{\"O} Arik and Tomas Pfister.
\newblock Tabnet: Attentive interpretable tabular learning.
\newblock In {\em Proceedings of the AAAI conference on artificial intelligence}, volume~35, pages 6679--6687, 2021.

\bibitem{baldi2015enhanced}
Pierre Baldi, Peter Sadowski, and Daniel Whiteson.
\newblock Enhanced higgs boson to $\tau$+ $\tau$- search with deep learning.
\newblock {\em Physical review letters}, 114(11):111801, 2015.

\bibitem{bertin2011million}
Thierry Bertin-Mahieux, Daniel~PW Ellis, Brian Whitman, and Paul Lamere.
\newblock The million song dataset.
\newblock 2011.

\bibitem{bishop1995neural}
Christopher~M Bishop.
\newblock {\em Neural networks for pattern recognition}.
\newblock Oxford university press, 1995.

\bibitem{blackard1999comparative}
Jock~A Blackard and Denis~J Dean.
\newblock Comparative accuracies of artificial neural networks and discriminant analysis in predicting forest cover types from cartographic variables.
\newblock {\em Computers and electronics in agriculture}, 24(3):131--151, 1999.

\bibitem{bommasani2021opportunities}
Rishi Bommasani, Drew~A Hudson, Ehsan Adeli, Russ Altman, Simran Arora, Sydney von Arx, Michael~S Bernstein, Jeannette Bohg, Antoine Bosselut, Emma Brunskill, et~al.
\newblock On the opportunities and risks of foundation models.
\newblock {\em arXiv preprint arXiv:2108.07258}, 2021.

\bibitem{borisov2022deep}
Vadim Borisov, Tobias Leemann, Kathrin Se{\ss}ler, Johannes Haug, Martin Pawelczyk, and Gjergji Kasneci.
\newblock Deep neural networks and tabular data: A survey.
\newblock {\em IEEE Transactions on Neural Networks and Learning Systems}, 2022.

\bibitem{sklearn_api}
Lars Buitinck, Gilles Louppe, Mathieu Blondel, Fabian Pedregosa, Andreas Mueller, Olivier Grisel, Vlad Niculae, Peter Prettenhofer, Alexandre Gramfort, Jaques Grobler, Robert Layton, Jake VanderPlas, Arnaud Joly, Brian Holt, and Ga{\"{e}}l Varoquaux.
\newblock {API} design for machine learning software: experiences from the scikit-learn project.
\newblock In {\em ECML PKDD Workshop: Languages for Data Mining and Machine Learning}, pages 108--122, 2013.

\bibitem{chang2023survey}
Yupeng Chang, Xu~Wang, Jindong Wang, Yuan Wu, Linyi Yang, Kaijie Zhu, Hao Chen, Xiaoyuan Yi, Cunxiang Wang, Yidong Wang, et~al.
\newblock A survey on evaluation of large language models.
\newblock {\em ACM Transactions on Intelligent Systems and Technology}, 2023.

\bibitem{chen2015xgboost}
Tianqi Chen, Tong He, Michael Benesty, Vadim Khotilovich, Yuan Tang, Hyunsu Cho, Kailong Chen, Rory Mitchell, Ignacio Cano, Tianyi Zhou, et~al.
\newblock Xgboost: extreme gradient boosting.
\newblock {\em R package version 0.4-2}, 1(4):1--4, 2015.

\bibitem{cortes1995support}
Corinna Cortes and Vladimir Vapnik.
\newblock Support-vector networks.
\newblock {\em Machine learning}, 20:273--297, 1995.

\bibitem{fayaz2022deep}
Sheikh~Amir Fayaz, Majid Zaman, Sameer Kaul, and Muheet~Ahmed Butt.
\newblock Is deep learning on tabular data enough? an assessment.
\newblock {\em International Journal of Advanced Computer Science and Applications}, 13(4):466--473, 2022.

\bibitem{friedman2001greedy}
Jerome~H Friedman.
\newblock Greedy function approximation: a gradient boosting machine.
\newblock {\em Annals of statistics}, pages 1189--1232, 2001.

\bibitem{geusebroek2005amsterdam}
Jan-Mark Geusebroek, Gertjan~J Burghouts, and Arnold~WM Smeulders.
\newblock The amsterdam library of object images.
\newblock {\em International Journal of Computer Vision}, 61:103--112, 2005.

\bibitem{gorishniy2022embeddings}
Yury Gorishniy, Ivan Rubachev, and Artem Babenko.
\newblock On embeddings for numerical features in tabular deep learning.
\newblock {\em Advances in Neural Information Processing Systems}, 35:24991--25004, 2022.

\bibitem{gorishniytabr}
Yury Gorishniy, Ivan Rubachev, Nikolay Kartashev, Daniil Shlenskii, Akim Kotelnikov, and Artem Babenko.
\newblock Tabr: Tabular deep learning meets nearest neighbors.

\bibitem{gorishniy2021revisiting}
Yury Gorishniy, Ivan Rubachev, Valentin Khrulkov, and Artem Babenko.
\newblock Revisiting deep learning models for tabular data.
\newblock {\em Advances in Neural Information Processing Systems}, 34:18932--18943, 2021.

\bibitem{grinsztajn2022tree}
L{\'e}o Grinsztajn, Edouard Oyallon, and Ga{\"e}l Varoquaux.
\newblock Why do tree-based models still outperform deep learning on typical tabular data?
\newblock {\em Advances in Neural Information Processing Systems}, 35:507--520, 2022.

\bibitem{guyon2019analysis}
Isabelle Guyon, Lisheng Sun-Hosoya, Marc Boull{\'e}, Hugo~Jair Escalante, Sergio Escalera, Zhengying Liu, Damir Jajetic, Bisakha Ray, Mehreen Saeed, Mich{\`e}le Sebag, et~al.
\newblock Analysis of the automl challenge series.
\newblock {\em Automated Machine Learning}, 177, 2019.

\bibitem{hegselmann2023tabllm}
Stefan Hegselmann, Alejandro Buendia, Hunter Lang, Monica Agrawal, Xiaoyi Jiang, and David Sontag.
\newblock Tabllm: Few-shot classification of tabular data with large language models.
\newblock In {\em International Conference on Artificial Intelligence and Statistics}, pages 5549--5581. PMLR, 2023.

\bibitem{hendrycks2016gaussian}
Dan Hendrycks and Kevin Gimpel.
\newblock Gaussian error linear units (gelus).
\newblock {\em arXiv preprint arXiv:1606.08415}, 2016.

\bibitem{hochreiter1997long}
Sepp Hochreiter and J{\"u}rgen Schmidhuber.
\newblock Long short-term memory.
\newblock {\em Neural computation}, 9(8):1735--1780, 1997.

\bibitem{huang2020tabtransformer}
Xin Huang, Ashish Khetan, Milan Cvitkovic, and Zohar Karnin.
\newblock Tabtransformer: Tabular data modeling using contextual embeddings.
\newblock {\em arXiv preprint arXiv:2012.06678}, 2020.

\bibitem{huang2012tabtransformer}
Xin Huang, Ashish Khetan, Milan Cvitkovic, and Zohar Karnin.
\newblock Tabtransformer: Tabular data modeling using contextual embeddings. arxiv 2020.
\newblock {\em arXiv preprint arXiv:2012.06678}, 2020.

\bibitem{ke2017lightgbm}
Guolin Ke, Qi~Meng, Thomas Finley, Taifeng Wang, Wei Chen, Weidong Ma, Qiwei Ye, and Tie-Yan Liu.
\newblock Lightgbm: A highly efficient gradient boosting decision tree.
\newblock {\em Advances in neural information processing systems}, 30, 2017.

\bibitem{ke2019deepgbm}
Guolin Ke, Zhenhui Xu, Jia Zhang, Jiang Bian, and Tie-Yan Liu.
\newblock Deepgbm: A deep learning framework distilled by gbdt for online prediction tasks.
\newblock In {\em Proceedings of the 25th ACM SIGKDD International Conference on Knowledge Discovery \& Data Mining}, pages 384--394, 2019.

\bibitem{ke2018tabnn}
Guolin Ke, Jia Zhang, Zhenhui Xu, Jiang Bian, and Tie-Yan Liu.
\newblock Tabnn: A universal neural network solution for tabular data.
\newblock 2018.

\bibitem{kohavi1996scaling}
Ron Kohavi et~al.
\newblock Scaling up the accuracy of naive-bayes classifiers: A decision-tree hybrid.
\newblock In {\em Kdd}, volume~96, pages 202--207, 1996.

\bibitem{lewis2020retrieval}
Patrick Lewis, Ethan Perez, Aleksandra Piktus, Fabio Petroni, Vladimir Karpukhin, Naman Goyal, Heinrich K{\"u}ttler, Mike Lewis, Wen-tau Yih, Tim Rockt{\"a}schel, et~al.
\newblock Retrieval-augmented generation for knowledge-intensive nlp tasks.
\newblock {\em Advances in Neural Information Processing Systems}, 33:9459--9474, 2020.

\bibitem{li2018massively}
Liam Li, Kevin Jamieson, Afshin Rostamizadeh, Ekaterina Gonina, Moritz Hardt, Ben Recht, and Ameet Talwalkar.
\newblock Massively parallel hyperparameter tuning.
\newblock 2018.

\bibitem{li2024cancergpt}
Tianhao Li, Sandesh Shetty, Advaith Kamath, Ajay Jaiswal, Xiaoqian Jiang, Ying Ding, and Yejin Kim.
\newblock Cancergpt for few shot drug pair synergy prediction using large pretrained language models.
\newblock {\em npj Digital Medicine}, 7(1):40, 2024.

\bibitem{liaw2018tune}
Richard Liaw, Eric Liang, Robert Nishihara, Philipp Moritz, Joseph~E Gonzalez, and Ion Stoica.
\newblock Tune: A research platform for distributed model selection and training.
\newblock {\em arXiv preprint arXiv:1807.05118}, 2018.

\bibitem{loshchilov2017decoupled}
Ilya Loshchilov and Frank Hutter.
\newblock Decoupled weight decay regularization.
\newblock {\em arXiv preprint arXiv:1711.05101}, 2017.

\bibitem{mcelfresh2024neural}
Duncan McElfresh, Sujay Khandagale, Jonathan Valverde, Vishak Prasad~C, Ganesh Ramakrishnan, Micah Goldblum, and Colin White.
\newblock When do neural nets outperform boosted trees on tabular data?
\newblock {\em Advances in Neural Information Processing Systems}, 36, 2024.

\bibitem{ozaki2020multiobjective}
Yoshihiko Ozaki, Yuki Tanigaki, Shuhei Watanabe, and Masaki Onishi.
\newblock Multiobjective tree-structured parzen estimator for computationally expensive optimization problems.
\newblock In {\em Proceedings of the 2020 genetic and evolutionary computation conference}, pages 533--541, 2020.

\bibitem{pace1997sparse}
R~Kelley Pace and Ronald Barry.
\newblock Sparse spatial autoregressions.
\newblock {\em Statistics \& Probability Letters}, 33(3):291--297, 1997.

\bibitem{NEURIPS2019_9015}
Adam Paszke, Sam Gross, Francisco Massa, Adam Lerer, James Bradbury, Gregory Chanan, Trevor Killeen, Zeming Lin, Natalia Gimelshein, Luca Antiga, Alban Desmaison, Andreas Kopf, Edward Yang, Zachary DeVito, Martin Raison, Alykhan Tejani, Sasank Chilamkurthy, Benoit Steiner, Lu~Fang, Junjie Bai, and Soumith Chintala.
\newblock Pytorch: An imperative style, high-performance deep learning library.
\newblock In {\em Advances in Neural Information Processing Systems 32}, pages 8024--8035. Curran Associates, Inc., 2019.

\bibitem{popov2019neural}
Sergei Popov, Stanislav Morozov, and Artem Babenko.
\newblock Neural oblivious decision ensembles for deep learning on tabular data.
\newblock {\em arXiv preprint arXiv:1909.06312}, 2019.

\bibitem{prokhorenkova2018catboost}
Liudmila Prokhorenkova, Gleb Gusev, Aleksandr Vorobev, Anna~Veronika Dorogush, and Andrey Gulin.
\newblock Catboost: unbiased boosting with categorical features.
\newblock {\em Advances in neural information processing systems}, 31, 2018.

\bibitem{quinlan1986induction}
J.~Ross Quinlan.
\newblock Induction of decision trees.
\newblock {\em Machine learning}, 1:81--106, 1986.

\bibitem{rubachev2022revisiting}
Ivan Rubachev, Artem Alekberov, Yury Gorishniy, and Artem Babenko.
\newblock Revisiting pretraining objectives for tabular deep learning.
\newblock {\em arXiv preprint arXiv:2207.03208}, 2022.

\bibitem{salzberg1994c4}
Steven~L Salzberg.
\newblock C4. 5: Programs for machine learning by j. ross quinlan. morgan kaufmann publishers, inc., 1993, 1994.

\bibitem{shwartz2022tabular}
Ravid Shwartz-Ziv and Amitai Armon.
\newblock Tabular data: Deep learning is not all you need.
\newblock {\em Information Fusion}, 81:84--90, 2022.

\bibitem{smith2019super}
Leslie~N Smith and Nicholay Topin.
\newblock Super-convergence: Very fast training of neural networks using large learning rates.
\newblock In {\em Artificial intelligence and machine learning for multi-domain operations applications}, volume 11006, pages 369--386. SPIE, 2019.

\bibitem{somepalli2021saint}
Gowthami Somepalli, Micah Goldblum, Avi Schwarzschild, C~Bayan Bruss, and Tom Goldstein.
\newblock Saint: Improved neural networks for tabular data via row attention and contrastive pre-training.
\newblock {\em arXiv preprint arXiv:2106.01342}, 2021.

\bibitem{vaswani2017attention}
Ashish Vaswani, Noam Shazeer, Niki Parmar, Jakob Uszkoreit, Llion Jones, Aidan~N Gomez, {\L}ukasz Kaiser, and Illia Polosukhin.
\newblock Attention is all you need.
\newblock {\em Advances in neural information processing systems}, 30, 2017.

\bibitem{wu2023brief}
Tianyu Wu, Shizhu He, Jingping Liu, Siqi Sun, Kang Liu, Qing-Long Han, and Yang Tang.
\newblock A brief overview of chatgpt: The history, status quo and potential future development.
\newblock {\em IEEE/CAA Journal of Automatica Sinica}, 10(5):1122--1136, 2023.

\bibitem{yang2023unitabe}
Yazheng Yang, Yuqi Wang, Guang Liu, Ledell Wu, and Qi~Liu.
\newblock Unitabe: A universal pretraining protocol for tabular foundation model in data science.
\newblock In {\em The Twelfth International Conference on Learning Representations}, 2023.

\end{thebibliography}

\end{document}